# Lung Cancer Concept Annotation from Spanish Clinical Narratives

Marjan Najafabadipour[1, [0000-0002-1428-9330]], Juan Manuel Tuñas[1[0000-0001-8241-5602]], Alejandro Rodríguez-González[1,2,* [0000-0001-8801-4762]] and Ernestina Menasalvas[1,2]

[1] Centro de Tecnología Biomédica, Universidad Politécnica de Madrid, Spain
[2] ETS de Ingenieros Informáticos, Universidad Politécnica de Madrid, Spain
[*] Corresponding author

m.najafabadipour@upm.es, juan.tunas@ctb.upm.es, alejandro.rg@upm.es, ernestina.menasalvas@upm.es

**Abstract.** Recent rapid increase in the generation of clinical data and rapid development of computational science make us able to extract new insights from massive datasets in healthcare industry. Oncological clinical notes are creating rich databases for documenting patient's history and they potentially contain lots of patterns that could help in better management of the disease. However, these patterns are locked within free text (unstructured) portions of clinical documents and consequence in limiting health professionals to extract useful information from them and to finally perform Query and Answering (Q&A) process in an accurate way. The Information Extraction (IE) process requires Natural Language Processing (NLP) techniques to assign semantics to these patterns. Therefore, in this paper, we analyze the design of annotators for specific lung cancer concepts that can be integrated over Apache Unstructured Information Management Architecture (UIMA) framework. In addition, we explain the details of generation and storage of annotation outcomes.

## 1 Introduction

Cancer is still one of the major public health issues, ranked with the second leading cause of death globally [1]. Across the Europe, lung cancer was estimated with 20.8% (over 266,000 persons) of all cancer deaths in 2011 [2] and the highest economic cost of 15% (18.8 billion) of overall cancer cost in 2009 [3]. Early diagnoses of cancer decreases its mortality rate [4]. Hence, a great attention on diagnoses is a key factor for both the effective control of the disease as well as the design of treatment plans.

Classically, the treatment decisions on lung cancer patients have been based upon histology of the tumor. According to World Health Organization (WHO), there are two broad histological subtypes of lung cancer: (1) Small Cell Lung Cancer (SCLC); and (2) Non-Small Cell Lung Cancer (NSCLC) [5]. NSCLC can be further defined at the molecular level by recurrent driver mutations [6] where mutations refer to any changes in the DNA sequence of a cell [7]. Tumor Mutations can occur in multiple oncogenes, including in: Epidermal Growth Factor Receptor (EGFR), Anaplastic Lymphoma Kinase (ALK), and Ros1 proto-oncogene receptor tyrosine kinase [8]. These oncogenes are Receptor Tyrosine Kinases, which can activate pathways associated with cell growth and proliferation [9]–[11].

One of the preliminary diagnoses factor of a cancer is its tumor stage. This factor plays a significant role on making decisions for developing treatment plans. The American Joint Committee on Cancer (AJCC) manual [13] specifies two standard systems for measuring the cancer stage [14]: (1) stage grouping and (2) TNM. The stage grouping system encodes the tumor stages using roman numerals, whereas the TNM system makes use of three parameters: (1) the size of tumor (T); (2) the number of lymph nodes (N); and (3) the presence of Metastasis (M).

According to International Consortium for Health Outcomes Measurement (ICHOM), Performance Status (PS) is a strong individual predictor of survival in lung cancer. The ICHOM working group recommended measuring PS as part of diagnoses per the Eastern Cooperative Oncology Group (ECOG) [14]. In addition to ECOG, Karnofsky is another scale for measuring PS [15]. These scales are used by doctors and researchers to assess the progress of a patient's disease, the effects of the disease on daily and living abilities of a patient and to determine appropriate treatment and prognosis [16].

Towards the digitization of medical data, these data have been stored in computerized medical records, named Electronic Health Records (EHRs). EHRs are rich clinical documents containing information about diagnoses, treatments, laboratory results, discharge summaries, to name a few, which can be used to support clinical decision support systems and allow clinical and translational research.

EHRs are mainly written mainly in textual format. They lack structure or have a structure depending on the hospital, service or even the physician generated them. They contain abbreviations and metrics and are written in the language of the country. Due to unstructured nature of information locked in EHRs, detection and extraction of useful information is still a challenge and consequences in difficulty of performing Q&A process [17].

To encode, structure and extract information from EHRs, an NLP system for which the Named Entity Recognition (NER) process is its paramount task, is required. Rule-based approaches for performing NER process through means of knowledge engineering are very accurate since they are based on physician's knowledge and experience [18].

The NER process intrinsically relies on ontologies, taxonomies and controlled vocabularies. Examples of such vocabularies are Systematized Nomenclature of Medicine (SNOMED) [19] and Unified Medical Language System (UMLS) [20]. The UMLS integrates and distributes key terminology, classification and coding standards. Even though that the translations of these vocabularies to different languages are available, they do not always provide the entire terminologies that are used in very specific domains (e.g., lung cancer). In addition, several medical metrics are not covered or fully provided by them. Furthermore, symbols such as "+" and "-", which are commonly being used with medical metrics (e.g., EGFR+) to determine their positivity or negativity, are not supported by them. Also, it is a common practice by physicians to use symbols such as ".", "_", "-", etc. for writing metrics (e.g., cancer stage I-A1). Such metrics are not supported by these ontologies as well.

Although, several NLP systems have been developed to extract information from clinical text such as Apache cTAKES [21], MEDLEE [22], MedTAS/P [23], HITEx [24], MetaMap [25], to name a few. However, these systems are mainly being used for English. One of the NLP systems that has been developed to perform IE on Spanish clinical text, is C-liKES (Clinical Knowledge Extraction System) [26]. C-liKES is a framework that has been developed on top of Apache UIMA, which has been based on a legacy system, named H2A [27].

To the best of our knowledge, there is no open NLP pipeline from which we can extract information related to lung cancer mutation status, tumor stage and PS, written in Spanish clinical narratives. Thus, the main contribution of this paper is to discuss the design, development and the implementation of annotators, capable of detecting clinical information from EHRs, using UIMA framework. Furthermore, we present the annotation results, extracted by means of running these annotators. The rest of paper is organized as follows: in Sect. 2, concept annotation for mutation status, stage and PS in lung cancer domain along with annotation output generation is presented; and in Sect. 3 the achievements gained so far are explained and the outlook of the future developments is provided.

## 2 Solution

We have developed a set of semantic rule-based NLP modules, named Annotators, using Apache UIMA framework. These annotators were developed to identify NEs (Named Entities) from clinical narratives. They contain regular expressions, using which, they can search for specific patterns through the clinical text.

The Pseudocode of algorithms implemented by these annotators is provided below.

```
Start;
  Set search-pattern = regex();
  Set input = EHR plain text;
  For sentences in input{
    For tokens in sentence{
      If search-pattern is matched to a token{
        Assign semantics to the matched token;
      }
      Move to the next token;
    }
    Move to the next sentence;
  }
End;
```

The algorithm defines the search pattern using regular expressions and accept EHR plain text as input. Then, for each individual sentence in the text, the algorithm checks if the search pattern can be matched with the tokens. Once, a matched token is found, the algorithm assigns semantical meanings to the concept.

To process a clinical narrative using these developed annotators, we have implemented them under a single pipeline. Once, the pipeline is executed, the output of annotations will be generated.

The outcomes of annotation processes are formatted as a set of XML Metadata Interchange (XMI) files and are also inserted into a relational database from which Q&A process can be followed. The details of lung cancer developed annotators and the output generation process are provided below.

### 2.1 Mutation Status

Physicians makes use of EGFR, ALK and ROS1 metrics for mentioning the tumor mutation status in clinical narratives. However, in case of EGFR, they can provide more detailed information about the mutation related to the exon (18–21), type of exon (deletion or insertion) and the mutation point (G719X, T790M, L858R, L861Q).

For determining the positivity or negativity of mutation metrics, physicians do not follow any standard systems. For example, in case of EGFR positive, , they can write: "*EGFR: positive"*, "*EGFR+*", "*has detected with mutation in EGFR*", "*presence of mutation in EGFR*", "*with insertion mutation in Exon 19*", "*EGFR mutated*", etc. Therefore, the need of annotators for detecting tumor mutation status from clinical text, comes to the picture. For this purpose, three annotators, named EGFR Annotator, ALK Annotator and ROS1 Annotator were developed using UIMA framework.

The EGFR annotator is capable of detecting the mutation status, exon, type of exon and the mutation point from the clinical text by incorporating four internal annotators that were developed for this purpose. Whereas the ALK and ROS1 annotators can only find the concepts that are related to the mutation status.

For example, in the following clinical text: "*EGFR + (del exon 19), no se detecta traslocación de ALK y ROS1 no traslocado.*" while EGFR mutation status is positive in exon 19, no translocation is detected for ALK and ROS1. Once, we process this clinical text using the developed annotators for tumor mutation status on UIMA CAS Visual Debugger (CVD), this information is extracted from the text (Fig. 1).

The CVD output representation is largely divided into two sections:

- Analysis Results: is composed of two subsections: (1) upper division: contains CAS Index Repository. *AnnotationIndex* represents the list of annotators executed for processing the text, which shows *DocumentAnnotation* objects, with one specific object for each specific annotation (ALK, EGFR, Exon, …). To see the annotation results of a specific annotator, the user should click on the designated one in here; and (2) lower division: includes AnnotationIndex. When the user clicks on the index of an annotation (in the figure ALK one) the information such as begin, end and semantic categories of the found NE will be represented.

- Text: accepts the clinical text as input from the user, which is in here “*EGFR + (del exon 19), no se detecta traslocación de ALK y ROS1 no traslocado.*”. The input text will be highlighted corresponding to the begin and end of annotation provided in the AnnotationIndex of the lower division sub-section of the Analysis Results section. The highlighted text is “*no se detecta traslocación de ALK”.*

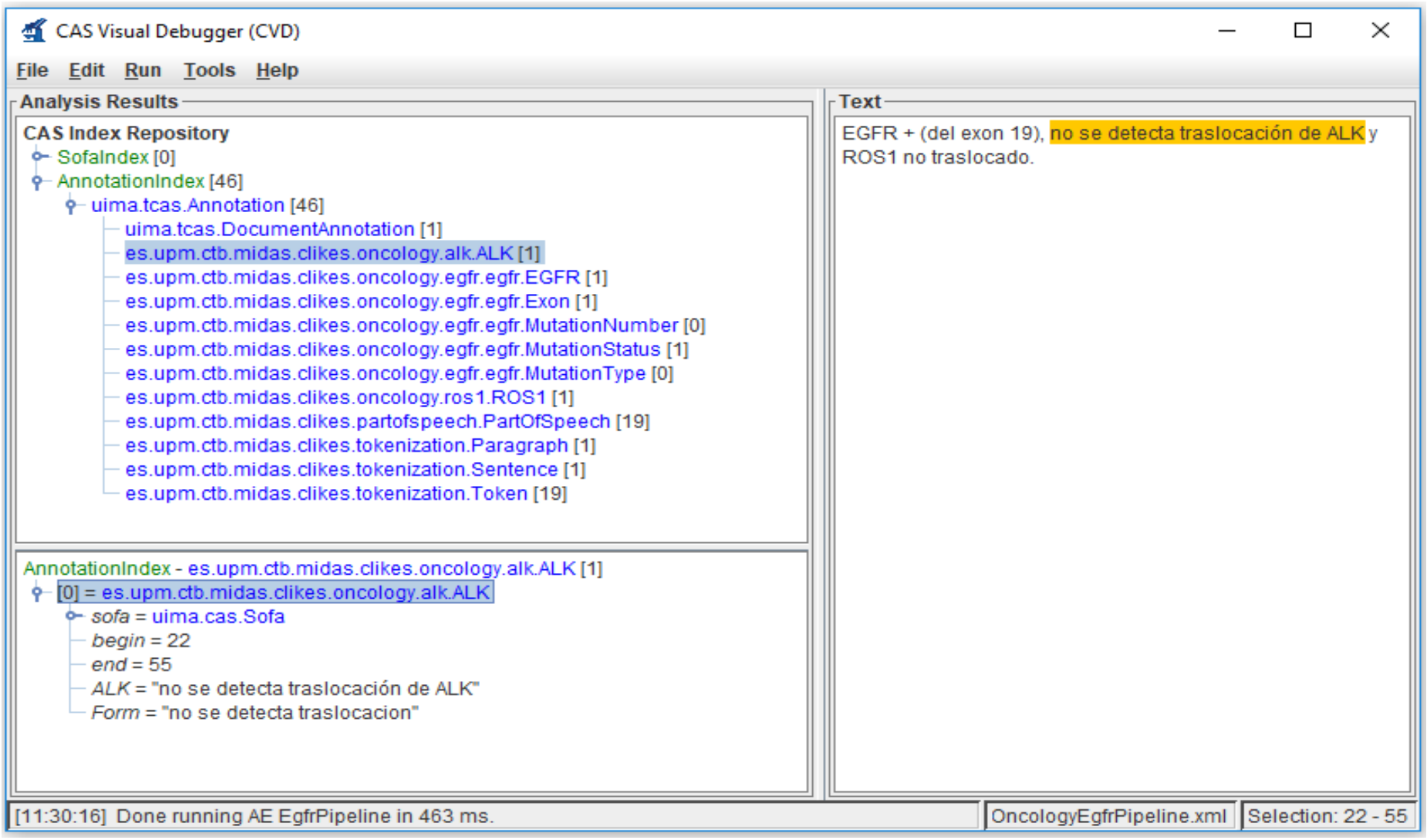

**Fig. 1.** EGFR, ALK and ROS1 annotation results on CVD

## 2.2 Stage

Stage grouping and TNM are the two main standard cancer staging systems, introduced by AJCC manual (Fig. 2). However, it is notable that for TNM system, the three attributes, i.e. T, N and M can be modulated by prefixes (e.g., pT1aN0M0). Such prefixes are: “c” (clinical), “p” (pathologic), “yc” or “yp” (post therapy), “r” (retreatment) and “a” (autopsy). Physicians normally use symbols such as “.”, “_”, “-”, “()”, etc. combined with TNM and stages metrics in the clinical text. For example, in case of stage IA1, they can write: *I-A1*, *I.A1*, *I_A_1*, *I(A1)*, etc.

| T/M | Subcategory | N0 | N1 | N2 | N3 |
|---|---|---|---|---|---|
| T1 | T1a | IA1 | IIB | IIIA | IIIB |
| | T1b | IA2 | IIB | IIIA | IIIB |
| | T1c | IA3 | IIB | IIIA | IIIB |
| T2 | T2a | IB | IIB | IIIA | IIIB |
| | T2b | IIA | IIB | IIIA | IIIB |
| T3 | T3 | IIB | IIIA | IIIB | IIIC |
| T4 | T4 | IIIA | IIIA | IIIB | IIIC |
| M1 | M1a | IVA | IVA | IVA | IVA |
| | M1b | IVA | IVA | IVA | IVA |
| | M1c | IVB | IVB | IVB | IVB |

**Fig. 2.** AJCC 8th edition - Lung cancer stage grouping and TNM system [28]

Using 8th edition of this manuals, we have developed two pattern-based extraction annotators, named Stage Annotator and TNM Annotator, for finding the cancer stages and the TNMs, appeared in the clinical narratives, respectively.

For example, the clinical text "*Adenocarcinoma de pulmón, pT1aN0M0 (micronódulos pulmonares bilaterales, linfangitis carcinomatosa, derrame pleural), estadio I_A1.*", explains that the patient is having lung cancer (Adenocarcinoma de pulmón) with pT1aN0M0 value for TNM and stage IA1. By processing this text using the Stage and TNM annotator, the information related to these two metrics has been annotated (Fig. 3).

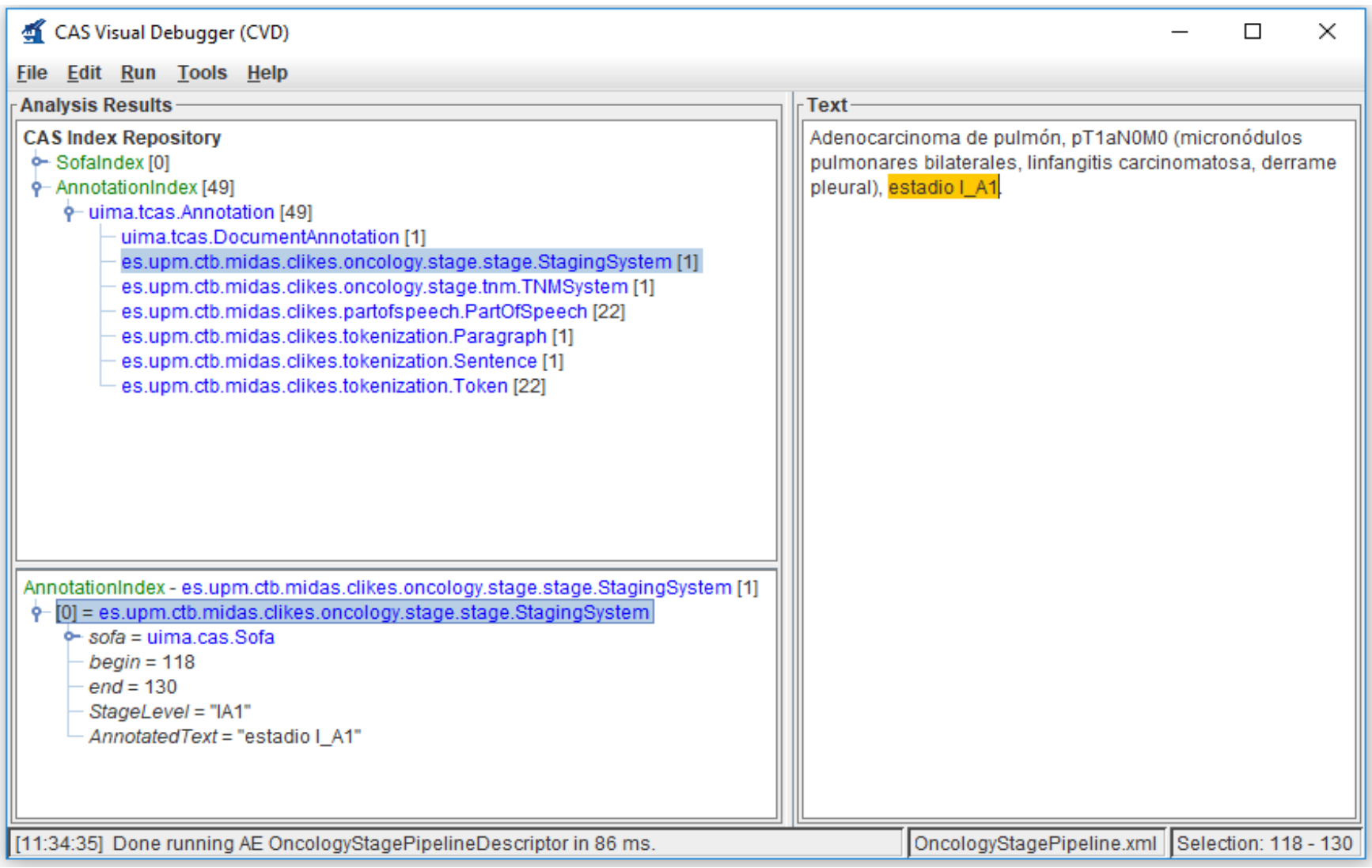

**Fig. 3.** Stage and TNM annotation results on CVD

## 2.3 PS

PS scale is mentioned using ECOG and Karnofsky measures in the clinical narratives. The ECOG measure ranges from 0 to 5, where 0 is the most ideal case for carrying on all pre-disease performance without any restrictions. On the other hand, the Karnofsky measure ranges from 0% to 100%, where 100% is the most ideal case.

ECOG and Karnfosky scales can appear with symbols such as ".", "_", "-", "()", etc. in clinical text. For example, ECOG 0 can be written as: "*ECOG_PS: 0*", "*ECOG-0*", "*ECOG is measured with 0*", "*ECOG (0)*", etc.

Hence, to annotate concepts related to ECOG and Karnofsky scales form clinical narratives, two annotators, named ECOG Annotator and Karnofsky Annotator were developed, respectively.

For example, "*ECOG-PS 0. Regular estado general. Disnea de reposo/mínimos esfuerzos. Karnofsky: 100%.*", indicates that the patient PS is measured with ECOG: 0 and Karnofsky: 100%. The results of annotation processes implemented by the ECOG and Karnofsky annotators on this clinical text, are shown in Fig. 4.

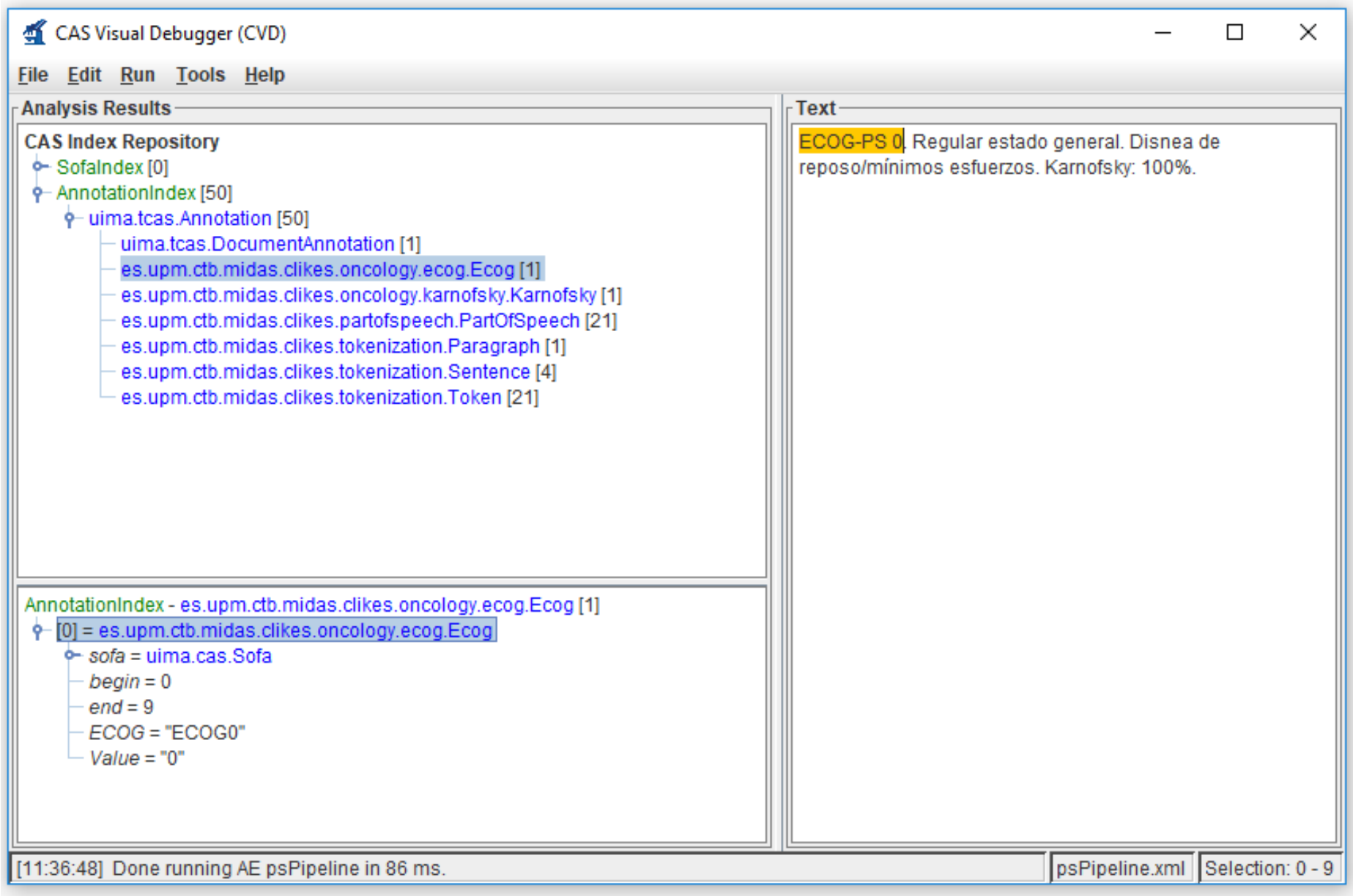

**Fig. 4.** ECOG and Karnofsky annotation results on CVD

## 2.4 Output Generation

For generating outcomes, an execution flow (Fig. 5) is followed by using a main process, called “Processing Engine”. This process accepts plain text files as input. Example of plain text documents are EHRs, clinical notes, radiology reports, and any kind of medical textual document generated.

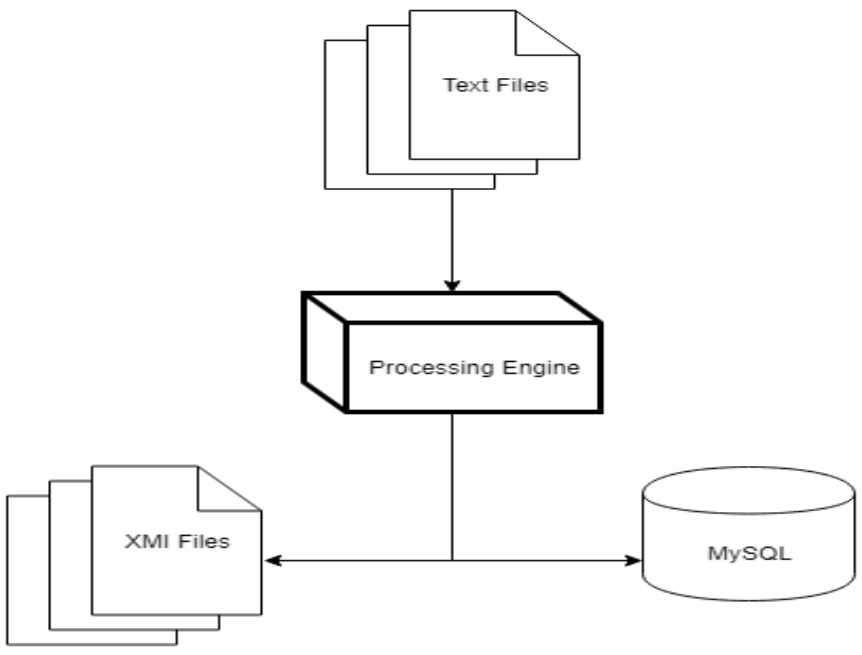

**Fig. 5.** Processing Engine Architecture

When the Processing Engine is executed, two resources are generated as output:

- XMI: UIMA annotators process plain text documents and generate one XMI file for each of them. These files encompass all the existing annotations i.e. they contain structured data of the relevant unstructured data in the EHR. Fig. 6 presents the results of annotations, stored in an XMI file, using UIMA Annotation Viewer. The UIMA annotation viewer is divided into three sections: (1) upper left division: contains plain clinical text. Highlighted tokens correspond to the annotated concepts, which are in here *"ECOG 3", "estadio IV", "EGFR", "no mutatdo" and "ALK no traslocado"*; (2) Annotation Type: provides the list of annotation tags from which the user can select the annotation results to be highlighted in the upper left division section. In here such tags are ALK, ECOG, EGFR, and etc.; and (3) Click In Text to See Annotation detail: by clicking on the highlighted tokens in the upper left division, the user can see the details of annotated concepts in this section. Such details are begin, end and semantic categories of the concept.
- Structured relational database: A MySQL database which contains the information of the annotations. The database allows to perform analysis on the structured data with more flexibility than XMI files.

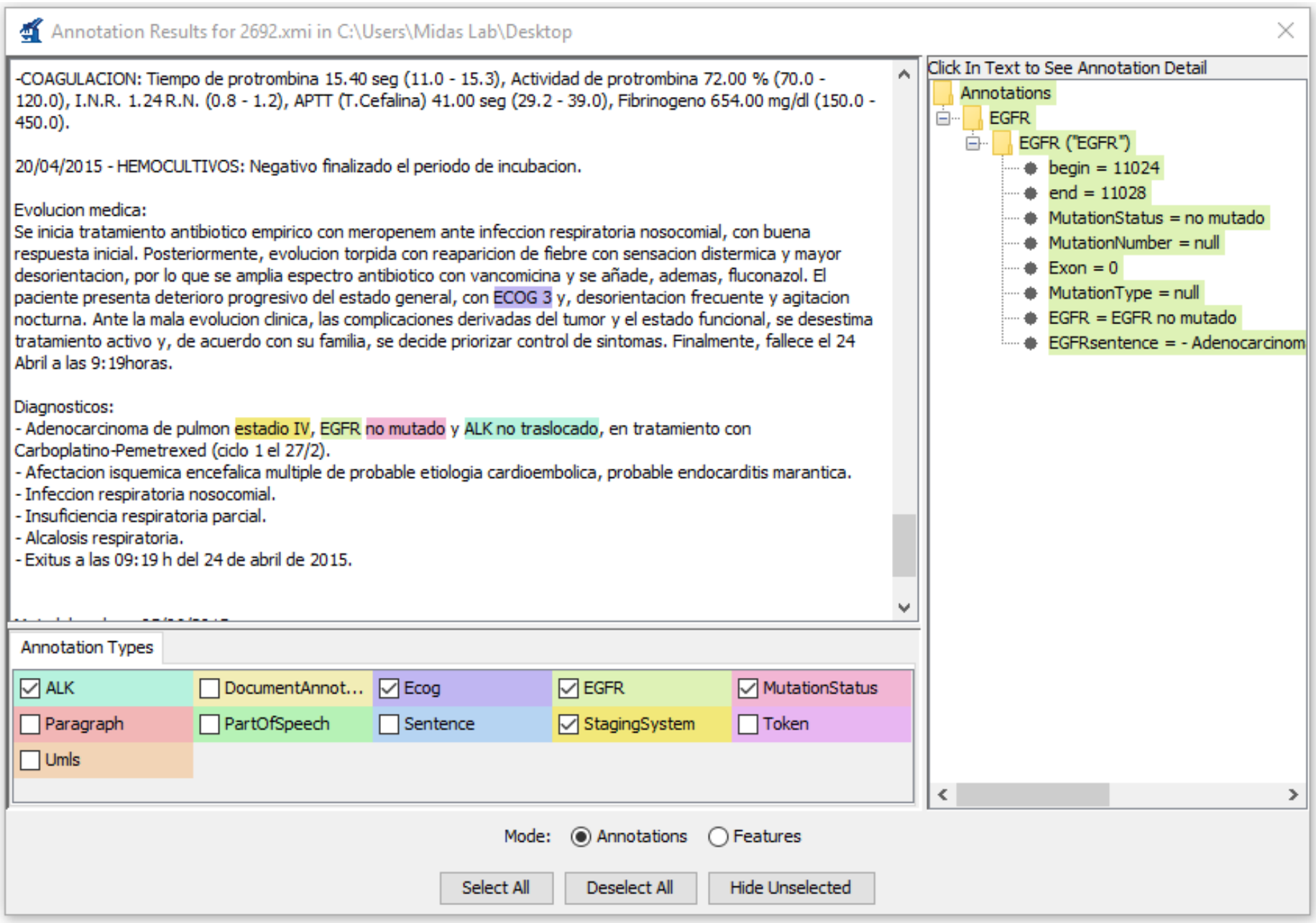

**Fig. 6.** XMI Annotation results using UIMA annotation viewer

## 3 Conclusion and Future Work

The vast amount of clinical data generated and the adaption of IT in health care industry, have motivated the development of NLP systems in clinical domain. For an NLP system to achieve a broad use, it must be capable of covering comprehensive clinical information and demonstrating effectiveness for a practical application. Thus, in this paper, we have described the development of specific case annotators for lung cancer domain, using UMIA framework. These annotators are capable of detecting information about tumor mutation status, stage of cancer and the PS from clinical text.

However, this work is an on-going research, which needs further improvements and developments. Such improvements will go into validations of already developed annotators whereas the developments will involve the semantic enrichment process for annotated medical concepts related to the lung cancer domain. Although, the recognition of medical concepts at NE level is one of the fundamental tasks of NLP but the judgment of clinical data cannot be understood solely at NE level. For example, clinicians can mention EGFR metric in the text for two reasons: (1) requesting for an EGFR test or (2) diagnoses of the cancer mutation status. To extract the patient's diagnosed mutation tumor status from clinical text, we need to have more semantic than NE level. Therefore, semantic enrichment process needs a great attention.